\documentclass[11pt]{article}
\usepackage[preprint]{acl}
\usepackage{times}
\usepackage{latexsym}
\usepackage[T1]{fontenc}
\usepackage[utf8]{inputenc}
\usepackage{microtype}
\usepackage{graphicx}
\usepackage{amsmath,amssymb}
\usepackage{booktabs,array,calc}
\providecommand{\tightlist}{\setlength{\itemsep}{0pt}\setlength{\parskip}{0pt}}

\title{What Does a Bias Head Measure? \\ Non-Identifiability in Head-Level Bias Attribution}
\author{S.M Seefat Alam \\
  Bangladesh University of Engineering and Technology, Dhaka \\
  \texttt{sakif.alam0015@gmail.com}}

\begin{document}
\maketitle
\begin{abstract}
Methods that localise social bias in large language models to individual
attention heads and remove it by ablation are validated by their effect on a
fairness metric, never by evidence that the statistic selecting the heads
measures bias. We audit this on Qwen-2.5-7B-Instruct and
Llama-3.1-8B-Instruct. The head map correlates at \(\rho = -0.950\) and
\(-0.768\) with a label-blind response-length map, apparently identifying a
surface confound. \textbf{An intervention that breaks the length--label
coupling rejects this reading:} with the prompt held fixed, response length
explains 1\% of the statistic's variance. What does explain it is prompt
identity, at 64\% of variance, in an ordering that is not monotonic in length.
Within any fixed prompt the fairness label, response length and the presence of
a protected-group token are collinear at \(\phi = 0.95\), partly by definition
of the labelling rule, so no observational analysis can attribute the map to
any of them, and no method in this family performs an experiment that could.
The substantive claims fail independently: bias selectivity is not
late-layer-concentrated on either model under either reference reading, and no
head set we tested reduces the unfairness rate when ablated. Its complement,
scored on the opposite label, reliably destroys refusal behaviour
(\(p = 2.7\times10^{-5}\)), demonstrating that the pipeline detects effects at
this scale.

\end{abstract}

\section{Introduction}\label{sec:introduction}\label{introduction}

A growing family of methods localises social bias in large language models to
individual attention heads and mitigates it by ablating them
\citep{han2026debiasing,jahan2026localizingrepairingbiastransformer,zayed2023fairnessawarestructuredpruningtransformers,islam2026biasgymsimplegeneralizableframework,yang2023biasahead}.
The statistics differ in kind but the structure is shared: a per-head score induces a ranking, a selection rule turns it into a set, and the set is zeroed. Most members compute their statistic on minimal pairs or at a single token position, where the two conditions contrasted are matched by construction. A newer design instead elicits free-form responses
under two prompting conditions, labels each fair or unfair with a judge, and
computes the statistic separately over each label set \citep{han2026debiasing}.
We audit that design in depth and ask a question that applies to the family as
a whole: what does a head-attribution statistic for social bias actually
measure, and what does ablating its output actually do?

We reproduce the paradigm on Qwen-2.5-7B-Instruct and Llama-3.1-8B-Instruct. The first result appears to be a clean answer. A control map built from the same per-example scores under an explicit alternative hypothesis --- response length, ignoring fairness labels entirely --- correlates at \(-0.950\) with the fairness map on Qwen and \(-0.768\) on Llama, with nine and eight of the ten top-ranked heads among each model's ten most length-suppressed. Four further controls agree, and the correlation \emph{strengthens} as the adversarial persona is weakened. A mechanism is available: committing to a group is terse and declining is verbose.

\textbf{Every piece of that evidence is correlational, and it is wrong.} We test this through an intervention. We ask the same questions using prompts that encourage either verbose or brief responses, filling the long-unfair and short-fair combinations missing from the original data. We keep the identified heads fixed throughout. Within a single
prompt the heads do not track length. In the terse condition, where responses average just ten tokens, length explains 1\% of the statistic's variance
while its correlation with the fairness label is \(+0.690\)
(Section~\ref{sec:testing-the-length-hypothesis-an-inter}). The pooled
correlation was a between-arm artifact. What dominates the score is prompt
identity: arm membership alone explains 64\% of variance, in an ordering not
monotonic in length. The construct itself proves unidentifiable --- within a
fixed prompt the fairness label and the presence of a protected-group token are
the same variable (\(\phi = 0.95\)), one direction of which is entailed by the
labelling rule. No observational analysis distinguishes them, and no method in
this family runs an experiment that would.

\textbf{Contributions.} (1) A correlational control that misleads and the
intervention that catches it
(Sections~\ref{sec:the-map-correlates-with-a-response-len}--\ref{sec:testing-the-length-hypothesis-an-inter}):
the diagnostic this literature lacks returned \(-0.950\) for a hypothesis an
intervention subsequently rejected, and every further correlational control
agreed with it. (2) Two non-reproductions with quantified uncertainty, and the
positive control that licenses them
(Sections~\ref{sec:layer-wise-localisation-does-not-repro}--\ref{sec:causal-evaluation}):
bias selectivity is not late-layer-concentrated on either model under either
reference reading, and no head set we tested reduces the unfairness rate when
ablated against a ten-draw layer-matched null, while its complement  reliably destroys refusal behaviour and replicates
across near-disjoint reselections (\(p = 1.75\times10^{-5}\)), the control
\citet{kendiukhov2026systematicevaluationsinglecellfoundation} identifies as
missing from ablation-based audits. (3) Two measurement hazards
(Sections~\ref{sec:causal-evaluation}--\ref{sec:ablation-degraded-generation-is-scored}):
a single random baseline sits inside a null band spanning 0.641--1.000, and
ablation-degraded generation is scored as fair, illustrated by a prescribed arm
that produced 31\% unclassifiable output and an apparent 0.127 reduction in
unfairness.

We reproduce the behavioural premise the paradigm rests on --- a DA versus CoT unfairness gap of 0.575 and 0.611 on the two models
(Section~\ref{sec:the-behavioural-premise-replicates}). What does not reproduce
is the mechanistic account of it.

\section{Related Work}\label{sec:related-work}\label{related-work}

\subsection{Head-level bias attribution}\label{sec:head-level-attribution-and-ablation-fo}

DiffHeads \citep{han2026debiasing} elicits responses under Direct-Answer (DA)
and Chain-of-Thought (CoT) prompting, labels each fair or unfair with an LLM
judge, computes a per-head contribution score over each label set, and ablates
heads scoring highly for unfair but not fair responses. The family shares that
structure and differs in the statistic --- minimal-pair scoring
\citep{jahan2026localizingrepairingbiastransformer}, leave-one-out ablation
\citep{zayed2023fairnessawarestructuredpruningtransformers}, first-token logit
attribution \citep{islam2026biasgymsimplegeneralizableframework}, head-mask
gradients \citep{yang2023biasahead}, annealing search
\citep{dasu2026attention} --- and
\citet{zhou2024unibiasunveilingmitigatingllm} applies it to in-context label
bias, a construct sharing only the word.

Validity checks are rare. \citet{yang2023biasahead} validate their map independently of downstream effect, mask the most negatively scored heads as an inverse control, and, recognising that selecting by the gradient of SEAT and evaluating with SEAT is circular, re-evaluate under an independent metric; \citet{islam2026biasgymsimplegeneralizableframework} probe one competing account qualitatively. Neither tests the map against a \emph{non-fairness} alternative, and later members inherit neither check. No paper reports any statistic of the generations its fairness metric scores, including their length. Random arms, where present, are a point estimate at one pruning ratio \citep{zayed2023fairnessawarestructuredpruningtransformers} or the best of several draws \citep{dasu2026attention}, never a null matched to the prescribed set; utility is measured off-distribution throughout. The source specification is itself loose --- reference-vector space
undefined, averaging set stated inconsistently, head-set equation contradicted
by its own figure caption, \(k\) unstated
(Appendix~\ref{app:deviations-from-the-source-specifi}) --- so we report both
reference readings throughout.

\subsection{Confounding, identifiability, and refusal}\label{sec:attribution-confounded-by-properties-o}\label{sec:identifiability-and-what-an-ablation-l}\label{sec:refusal-and-what-the-fairness-label-is}

That an interpretability result can track something other than the concept it names is an established failure mode \citep{adi2016fine,hewitt2019designing,kumar2022probing,adebayo2018sanity,adebayo2022post}. \textbf{The closest prior result is \citet{tan2024analysing}}: contrastively extracted steering vectors have highly variable per-input effects, much of that variance a steerability bias tied to answer position and token identity --- the prompt template rather than the target behaviour. We find a contrastive \emph{head-attribution} statistic likewise dominated by prompt identity. \textbf{A second neighbour is instructive for a result we did not obtain:} \citet{dahlem2026selfattentiontransportlimitssymmetric} show spectral diagnostics of the attention operator inheriting a response-length confound in hallucination detection. We expected the same here; the correlational evidence was overwhelming and an intervention rejected it (Sections~\ref{sec:the-map-correlates-with-a-response-len}--\ref{sec:testing-the-length-hypothesis-an-inter}).

\citet{meloux2025everything} argue that mechanistic explanations are frequently non-identifiable, and \citet{meloux2025dead} show pipelines returning plausible explanations for randomly initialised networks, concluding that findings must be tested against explicit alternatives with quantified uncertainty; this paper is an instantiation. \citet{meloux2025mechanistic} give the statistical form of our null band \citep[see also][]{quirke2026ablation,xu2026closurevalidatedcircuitdiscoveryattention}, and that most heads prune without measurable loss \citep{voita2019analyzing} makes an unmatched single random baseline uninformative. The closest methodological precedent is \citet{kendiukhov2026systematicevaluationsinglecellfoundation}, in genomics: top-ranked ``regulatory'' heads degrade nothing at any dose while twenty random heads do, with an intervention-fidelity check before the null is interpreted --- a diagnostic we adopt (Section~\ref{sec:intervention}) --- and its stated limitation is ours to address, that no positive control shows the pipeline can recover known causal structure. Our complement set is one: not the bottom of the ranking \citep[cf.][]{yang2023biasahead} but the top computed on the \emph{opposite label set}. The construction is not exotic --- \citet{islam2026biasgymsimplegeneralizableframework} already partition heads by score sign and ablate one side --- so the family has the vocabulary for this control and has not run it.

That our reverse set destroys refusal places it in mapped territory: refusal is mediated by an ablatable direction \citep{arditi2024refusal} localised to small head sets \citep{rdsha2025,refusalcliff2025}, and \citet{zhao2025chain} select heads by the association between CoT \emph{length} and attention ratio and show that ablating them weakens refusal. We claim its use as a validity argument, not the phenomenon: because this paradigm defines a fair response as a refusal \citep{han2026debiasing}, a statistic recovering causal heads on the refusal side and nothing on the committal side is a statement about the construct. \citet{blodgett2020,blodgett2021} argue that bias measurements
frequently lack an articulated construct; we apply that critique one level
down, to the attribution built on such a measurement \citep{cronbach1955}. CoT
does not reliably reduce social bias \citep{shaikh2023}, and persona
instructions substantially shift it \citep{personasurvey2024} --- our
manipulation check finds the persona worth 0.787 on the outcome measure
(Section~\ref{sec:the-outcome-measure-is-not-saturated}).

\section{Method}\label{sec:method}\label{method}

\subsection{Setup}\label{sec:what-we-audit}\label{sec:models}\label{sec:elicitation}\label{sec:response-labelling-and-validation}

We audit contrastive head-level fairness attribution: elicit responses under
two prompting conditions, label them fair/unfair, compute a per-head score over
each label set, and ablate heads high for unfair but not fair. DiffHeads
\citep{han2026debiasing} is the instantiation we reproduce.

\textbf{Models.} Qwen-2.5-7B-Instruct \citep{qwen2025} (28 layers $\times$ 28 heads) and Llama-3.1-8B-Instruct \citep{llama3} (32 layers, 32 query / 8 KV heads), both \(d_h = 128\), run in 4-bit NF4 \citep{dettmers2023qlora} under Hugging Face Transformers \citep{wolf2020transformers}, greedy decoding. All prompts and generations are in English. All head-level operations index the input to \texttt{o\_proj} by \emph{query} head; grouped-query attention affects only the key--value pathway, and \texttt{o\_proj} carries no bias, so zeroing an input slice exactly subtracts that head's contribution.

\textbf{Elicitation.} Role-play forced choice between two social groups on a trait (``Who do you think is \emph{more trustworthy}? \emph{Men} or \emph{women}?'') under a persona, in two styles: \textbf{DA} (no reasoning) and \textbf{CoT} (``Let's think step by step''). Identification uses 60 questions $\times$ both styles at \texttt{max\_new\ =\ 128}, regenerated under three persona strengths (full, mild, none) because the persona strongly affects the outcome; evaluation uses 80 balanced held-out questions (8 protected pairs $\times$ 10) with no trait--pair overlap. Two earlier evaluation sets (Appendix~\ref{app:additional-evaluation-sets-and-ful}) are used only for the degradation analysis.

\textbf{Labelling.} A response is \emph{unfair} if it commits to one of the two groups, \emph{fair} if it declines or hedges, and unclassifiable otherwise. The labeller inspects the opening 250 characters, with three negation-aware rules added after an audit of all stored responses (Appendix~\ref{app:labeller-development-failure-audit}). Unlike the original, which uses an LLM judge without reported validation, we validate against human annotation \textbf{on the evaluation data itself} --- 120 responses drawn blind and stratified across all 26 arm strata of both models' evaluation files --- so the validation applies to the results rather than to a predecessor instrument (Table~\ref{tab:labeller}). Agreement is 0.907 within the labeller's fixed reading window and 0.696 beyond it, which bears asymmetrically on the arms (Limitations).

\subsection{Score, selection, and intervention}\label{sec:contribution-score}\label{sec:head-selection-and-controls}\label{sec:intervention}

For response tokens \(R\), head output \(z^{(l)}_{r,h}\) (the
\texttt{o\_proj} input slice), output-projection block \(W^{(l)}_{O,h}\), and
unit reference \(\bar v_{\text{ref}}\):
\[S^{(l)}_h = \frac{1}{|R|}\sum_{r \in R}\Big[\big(W^{(l)}_{O,h} z^{(l)}_{r,h}\big)^{\!\top}\bar v_{\text{ref}}\Big]_+^2,\]
computed separately over fair and unfair sets and standardised \textbf{within
each layer} (residual-stream norms grow with depth; flattened standardisation
ranks layers, not heads). The original underspecifies \(\bar v_{\text{ref}}\)'s
space; we implement final-layer and layer-matched readings, report
layer-matched as primary (higher reliability on both models), and verify every
conclusion under both. Scores use at most the first 64 response tokens; a fixed 12-token window control is reported in Appendix~\ref{app:further-correlational-controls}.

With \(\mathcal{T}_K\) the top-\(K\) heads, the prescribed \emph{differential-unfair} set is \(\mathcal{H}_{\text{diff}} = \mathcal{T}_K(\tilde S^{\text{unfair}}) \setminus \mathcal{T}_K(\tilde S^{\text{fair}})\), the set difference being the only reading consistent with the name (the source's Equation 4 contradicts its own figure caption here, Appendix~\ref{app:deviations-from-the-source-specifi}); \(k\) is unstated and we set \(K = 20\), which yields 10 heads on both models. The \textbf{reverse set}, the same difference taken the other way and truncated to equal cardinality, is our positive control. \textbf{Layer-matched random sets} replace each prescribed head with another from the same layer, ten draws per model, because the null is wide and an unmatched or single control confounds depth with identity. Selection provenance is in Appendix~\ref{app:selection-provenance} and the head-set invariance under the labeller correction in Appendix~\ref{app:head-selection-invariance}.

\textbf{Masking} zeroes each selected head's contribution at \textbf{every} decoding position via a forward hook on \texttt{o\_proj}; masking only the final prompt position removes almost nothing. Before every masking run we require next-token \(\text{KL}(\text{base}\,\|\,\text{masked}) > 10^{-4}\) and differing greedy text (observed 0.006--0.211 on Qwen, 0.075 on Llama) --- adopted after an early run with silently detached hooks produced statistics indistinguishable from a genuine null.

\subsection{Evaluation and statistics}\label{sec:evaluation-and-control-analyses}\label{sec:statistical-analysis}\label{sec:deviations-from-the-source}

Primary outcome: unfair / (unfair + fair), with unclassifiable responses
reported separately and never folded into the denominator --- folding them in
scores breakage as improvement (Section~\ref{sec:causal-evaluation}). Controls: split-half \textbf{reliability} of every map (Spearman--Brown corrected), a label-permutation \textbf{placebo}, a five-signal \textbf{degeneracy screen}, a prompt-only \textbf{manipulation check}, and a three-strength \textbf{persona ablation}. Wilson intervals; Fisher's exact for pairs; McNemar item-level with Holm correction across arm families; permutation nulls for layer distributions; Spearman over all \(L \times H\) heads, disattenuated by \(\sqrt{\rho_1\rho_2}\). \emph{Holm correction is applied within each model
across all clean arms compared to baseline (11 arms on Qwen, 10 on Llama),
using the exact two-sided binomial on discordant pairs with monotonicity
enforced; different models are different families, and the reselection arms of
Section~\ref{sec:causal-evaluation} form their own family of two.}

We deviate from the source in four respects of scale and instrumentation --- 60 identification questions against 2,516, a validated rule-based labeller rather than an LLM judge, a purpose-built question set, and 4-bit quantisation --- and three concerning the statistic: a 64-token cap on the averaging set, the reading of \(z^{(l)}_{r,h}\) as the per-head slice of the attention output rather than the value vector, and the reference-vector reading above. Appendix~\ref{app:deviations-from-the-source-specifi} gives each in full and the Limitations section states what they license.

\begin{table*}[t]
\centering
\small
\begin{tabular}{@{}lll@{}}
\toprule
& primary labeller & predecessor \\
\midrule
agreement & \textbf{0.867} {[}0.794, 0.916{]} & 0.842 {[}0.766, 0.896{]} \\
false-unfair rate & \textbf{0.120} {[}0.042, 0.300{]} & 0.160 {[}0.064, 0.347{]} \\
false-fair rate & 0.044 {[}0.017, 0.108{]} & 0.044 {[}0.017, 0.108{]} \\
unclassifiable & 11/120 & 13/120 \\
\bottomrule
\end{tabular}
\caption{Labeller validation against blind human annotation drawn from the evaluation data itself. Wilson intervals; directional error rates are conditioned on the human label. Agreement does not differ detectably across arm types (Appendix~\ref{app:labeller-development-failure-audit}).}\label{tab:labeller}
\end{table*}

\section{Results}\label{sec:results}\label{results}

\subsection{Determinism and intervention
fidelity}\label{sec:determinism-and-intervention-fidelity}\label{determinism-and-intervention-fidelity}

24/24 byte-identical responses across independent runs; masking KL
0.006--0.211 (Qwen), 0.075 (Llama) with altered greedy text.

\subsection{The behavioural premise
replicates}\label{sec:the-behavioural-premise-replicates}\label{the-behavioural-premise-replicates}

The unfairness rate is 0.800 under DA against 0.225 under CoT on Qwen, and 0.778 against 0.167 on Llama --- gaps of 0.575 and 0.611 (Appendix~\ref{app:unfairness-by-elicitation}). Everything that follows concerns the mechanistic account, not the phenomenon.

\subsection{The map correlates with a response-length direction}\label{sec:the-map-correlates-with-a-response-len}\label{the-map-correlates-with-a-response-length-direction}

We build a \textbf{length map} from the same per-example scores, contrasting responses above and below median length while ignoring fairness labels. If the statistic measures bias the fairness map should be distinguishable from it. It is not: the two correlate at \(\rho = -0.950\) on Qwen and \(-0.768\) on Llama, and nine of the ten highest-scoring heads in the fairness map are among the ten most length-suppressed. The maps are reliably estimated (split-half 0.950 and 0.959, collapsing to approximately zero under label permutation), so this is not sampling noise. Four further controls all agree with the length reading, and the association \emph{strengthens} as the adversarial persona is weakened (Appendix~\ref{app:further-correlational-controls}). Every result here is correlational, on data where label and length are close to inseparable. \textbf{Section~\ref{sec:testing-the-length-hypothesis-an-inter} tests the hypothesis directly and rejects it.}

\subsection{Testing the length hypothesis: an intervention}\label{sec:testing-the-length-hypothesis-an-inter}\label{testing-the-length-hypothesis-an-intervention}

The evidence above cannot distinguish a head map that tracks response length
from one that tracks the fairness label, because in the identification data the
two are confounded at AUROC 0.878--0.996. We break the coupling by instruction
and ask which the statistic follows.

\textbf{Design.} The same 60 questions, persona, model and greedy decoding under four prompts: the existing DA and CoT conditions plus \emph{verbose} (state the choice, then explain in detail) and \emph{terse} (answer in one short sentence) instructions, none mentioning fairness, stereotypes or groups. The two new arms populate the cells the original data lacks, long-unfair and short-fair. The ten prescribed heads are loaded fixed; nothing is re-identified. Of 240 generations, 220 receive a label (151 unfair, 69 fair).

\textbf{Pooled, the length hypothesis appears confirmed} (\(\rho = -0.758\), \(p = 2.3\times10^{-42}\)). \textbf{Within arms, with the prompt held constant, it reverses.}

\begin{table}[t]
\centering
\small
\begin{tabular}{@{}llll@{}}
\toprule
arm & \(n\) & \(r\)(score, length) & \(r\)(score, unfair) \\
\midrule
DA & 60 & -0.679 & \textbf{+0.780} \\
CoT & 40 & \textbf{+0.224} & \textbf{+0.615} \\
verbose & 60 & -0.277 & \textbf{+0.387} \\
terse & 60 & \textbf{-0.025} & \textbf{+0.725} \\
\bottomrule
\end{tabular}
\caption{Correlation of the head score with response length and with the fairness label, within each elicitation arm.}\label{tab:withinarm}
\end{table}

The label correlation is strong and identically signed in all four arms; the length correlation is strongly negative in one, near zero in another and \textbf{positive} in a third. The pooled figure is a between-arm effect: the arms differ in mean length by roughly 38$\times$ and also in prompt text.

\textbf{The terse arm is decisive.} Its responses average 10.1 tokens (sd 5.6), so a length direction could not vary much within it. Instead \textbf{length explains 1.0\% of score variance} (\(\rho = -0.025\), \(p = 0.85\)) while the label correlation is \(+0.690\) (\(p = 1.1\times10^{-9}\)). \textbf{The length hypothesis is rejected.}

\textbf{What the score tracks is not the fairness construct, distinguishably.} Within the terse arm the fairness label and the presence of a protected-group token are effectively the same variable (\(\phi = 0.951\); 59 of 60 concordant). One direction is definitional, since the labeller returns unfair only via group-name matching; the empirical content is the other cell, where \textbf{12 of 13 fair responses contain no protected-group token at all} because terse refusals negate at the trait level (``not determined by gender'') rather than naming the groups. Variance explained is 0.010 for length alone, 0.534 adding group-token, 0.550 adding the label instead, 0.558 for all three; the increments between the label and group-token models are not interpretable at this collinearity. The two hypotheses are not separable under this operationalisation --- a statement about the labelling rule, not about the heads.

\textbf{Prompt identity dominates.} Arm membership alone explains \(R^2 = 0.635\) of score variance, more than length (0.433) and the label (0.273) combined, and the per-arm partial effects are not monotonic in length: terse (\textasciitilde10 tokens) \(+0.538\) and CoT (\textasciitilde384 tokens) \(+0.479\) are both strongly positive while verbose (\textasciitilde165 tokens) is \(-0.641\). A length direction cannot produce that ordering.

\textbf{Prediction accountability.} We predicted before running this experiment
that the prescribed heads would track length: that \(|\rho(\text{score},
\text{length})| > 0.5\), that the point-biserial correlation with the label
would be near zero, that the length partial would survive controlling for the
label, and that the label partial would not. Two of the four held. The two
concerning the label failed. We report the prediction and its failure because
the correlational evidence in
Section~\ref{sec:the-map-correlates-with-a-response-len} is exactly the
evidence that would have supported it, and the discipline this paper recommends
(Section~\ref{sec:what-a-method-in-this-family-should-re}) is the reason the
failure was detected rather than assumed away.

\subsection{Layer-wise localisation does not reproduce}\label{sec:layer-wise-localisation-does-not-repro}\label{layer-wise-localisation-does-not-reproduce}

Under the layer-matched reference, bias selectivity sits significantly \textbf{earlier} than chance on both models (centre-of-mass 11.72 against a uniform null of 13.5 on Qwen, \(p = 0.011\); 13.67 against 15.5 on Llama, \(p = 0.021\)), and the final four layers carry under half their chance share of selectivity. That \emph{early} concentration is not invariant to the reference reading, but the negative claim is: of the eight model \(\times\) reference \(\times\) labeller combinations we computed, seven place selectivity earlier than chance and \textbf{none shows the late concentration the method reports}. With map reliability above 0.95 this is not estimation noise (Appendix~\ref{app:layer-distribution}).

\subsection{Causal evaluation}\label{sec:causal-evaluation}\label{causal-evaluation}

Thirteen arms per model on 80 held-out balanced questions --- baseline (rates 0.762 on Qwen, 0.708 on Llama), the prescribed set, its complement, and ten layer-matched random draws --- plus, because the two reference readings select substantially different heads (2 of 10 shared), the prescribed set and its complement under the reading each model's stored selection did \emph{not} use.

\subsubsection{The prescribed set}\label{sec:the-prescribed-set}\label{the-prescribed-set}

\begin{table*}[t]
\centering
\small
\begin{tabular}{@{}llllllll@{}}
\toprule
model & selection & rate & $\Delta$ & u$\rightarrow$f / f$\rightarrow$u & Holm & degen / unclass. & tokens \\
\midrule
Qwen & final-layer (stored) & 0.775 & +0.013 & 1 / 2 & 1.000 & 0 / 0 & 21.4 $\rightarrow$ 47.4 \\
Qwen & \textbf{layer-matched} & 0.873 & \textbf{+0.111} & \textbf{0 / 8} & \textbf{0.0156} & 0 / 1 & --- \\
Llama & layer-matched (stored) & 0.795 & +0.086 & 3 / 11 & 0.344 & 3 / 7 & 27.2 $\rightarrow$ 60.0 \\
Llama & final-layer & 0.582 & -0.127 & --- & --- & 1 / \textbf{25} & --- \\
\bottomrule
\end{tabular}
\caption{Prescribed set against baseline. $\Delta$ is the change in unfairness rate; transitions are item-level over paired questions.}\label{tab:prescribed}
\end{table*}

Llama's final-layer arm returned 31\% unclassifiable output, exceeding the 20\% threshold applied throughout, and is \textbf{excluded}; its apparent 0.127 reduction is breakage, not debiasing (Section~\ref{sec:ablation-degraded-generation-is-scored}).

\textbf{The null is selection-specific.} For the stored Qwen set masking changes almost nothing ($\Delta$ = +0.013, one flip each way, Holm 1.000); for the set selected under the primary reading the effect is significant at the item level (0 unfair$\rightarrow$fair against 8 fair$\rightarrow$unfair, \(p = 0.0078\)). ``Masking the prescribed heads has no measurable effect'' holds for one selection and fails for another.

\textbf{No selection reduces unfairness.} In every evaluable arm --- four prescribed, three complement, two models, two reference readings --- item-level movement is net \emph{toward} unfairness (+1, +8, +8 prescribed; +8, +7, +13, +16 complement). Not one moves in the direction the method claims, while layer-matched random draws move refusal in both directions: the consistency is a property of the selected sets, not of ablation in general. On both models masking more than doubles response length without a corresponding change in verdicts.

Masking's effect on response length also interacts with the labeller's reading window; correcting for it \textbf{increases} \(\Delta\) under every error model examined, so the reported value is a lower bound (Appendix~\ref{app:measurement-exposure}).

\subsubsection{The complement set}\label{sec:the-complement-set}\label{the-complement-set}

\begin{table}[t]
\centering
\footnotesize
\begin{tabular}{@{}llllll@{}}
\toprule
model & sel. & lab. & u\,$\to$\,f & f\,$\to$\,u & \(p\) \\
\midrule
Qwen & stored & prim. & \textbf{0} & 8 & \(7.8\times10^{-3}\) \\
Qwen & stored & prev. & \textbf{0} & 9 & \(3.9\times10^{-3}\) \\
Llama & stored & prim. & \textbf{0} & 13 & \(2.4\times10^{-4}\) \\
Llama & stored & prev. & \textbf{0} & 13 & \(2.4\times10^{-4}\) \\
\midrule
Qwen & resel. & prim. & 1 & 8 & \(3.9\times10^{-2}\) \\
Qwen & resel. & prev. & \textbf{0} & 8 & \(7.8\times10^{-3}\) \\
Llama & resel. & prim. & \textbf{0} & 16 & \(3.1\times10^{-5}\) \\
Llama & resel. & prev. & \textbf{0} & 16 & \(3.1\times10^{-5}\) \\
\bottomrule
\end{tabular}
\caption{Complement-set item-level transitions; \emph{prim.}\ and \emph{prev.}\ are the primary and predecessor labellers. Rows above the rule are the stored selections; rows below are the reselections under the alternative reference reading, whose head sets share 2--3 of 10 heads with them.}\label{tab:complement}
\end{table}

Fisher combination across the two models under the primary labeller is \(\chi^2 = 26.34\), \(p = 2.7\times10^{-5}\) for the stored selections and \(\chi^2 = 27.28\), \(p = 1.75\times10^{-5}\) for the reselections, df = 4 (only across-model pairs are combined; the labeller variants score the same generations). \textbf{One counter-directional flip appears in the entire study}, in the reselected Qwen arm under the primary labeller; every stored arm is zero under both labellers. The effect survives reselection onto a near-disjoint head set on both models, though the \emph{asymmetry} between the prescribed set and its complement does not (Section~\ref{sec:what-the-complement-set-shows}).

\subsubsection{The random null}\label{sec:the-random-null}\label{the-random-null}

The ten-draw layer-matched band spans 0.641--1.000 on Qwen (9 clean draws) and 0.603--0.886 on Llama (8 clean). The stored prescribed sets sit at the 33rd and 62nd percentiles (\(z = -0.36\), \(+0.62\)), the stored complement sets at \(z = +0.37\) and \(+2.13\), the latter outside its band; rate-level evidence for the complement effect is therefore mixed across models and the item-level asymmetry in Table~\ref{tab:complement} carries the claim. The band's width is itself a finding: a single random control, standard in this literature, lands anywhere within it.

\emph{The band is layer-matched to the stored head sets, so it does not transfer to Qwen's reselected set, whose layer profile is substantially earlier (mean layer 11.5 against 15.8); that set's percentile position is therefore omitted and only its item-level result is reported.}

\subsection{Ablation-degraded generation is scored as
fair}\label{sec:ablation-degraded-generation-is-scored}\label{ablation-degraded-generation-is-scored-as-fair}

\emph{Per-arm degeneracy and unclassifiable counts are in Table~\ref{tab:prescribed} and Appendix~\ref{app:full-degradation-analysis-ev-2-qwe}.}

The clearest instance arises in a prescribed arm. Masking Llama's heads as selected under the final-layer reference produced 31\% unclassifiable output, the highest next-token KL of any arm (0.386), and an apparent \textbf{0.127 reduction} in the unfairness rate --- the only movement in the method's claimed direction anywhere in our results, and breakage rather than debiasing. This is the hazard in its consequential form: the intervention a practitioner would deploy, producing a number that looks like successful debiasing.

Validity destruction is heterogeneous across draws: unclassifiable counts range
from 1 to 54 of 80 on Qwen and 4 to 23 on Llama. The denominator convention
matters as much as the screen that detects it: scoring unclassifiable responses
as ``not unfair'' credits breakage as improvement, and did so in an earlier
version of our own analysis before it was caught
(Appendix~\ref{app:reproducibility-notes}). The screen's five signals, its known
insensitivity to typo-level corruption, and a second instance of the hazard on
an earlier evaluation set are in
Appendix~\ref{app:degeneracy-and-the-denominator}.

\subsection{Outcome measure and labeller invariance}\label{sec:the-outcome-measure-is-not-saturated}\label{the-outcome-measure-is-not-saturated}

\textbf{Prompt conditions without masking} (Qwen, 61 protected questions, no masking): full
persona + DA 0.967; no persona \textbf{0.180} (\(\Delta=-0.787\),
\(p<10^{-4}\)); explicit fairness instruction \textbf{0.279}
(\(\Delta=-0.689\), \(p<10^{-4}\)); zero unclassifiable or degenerate
responses in either condition. The prompt moves the metric by 0.787;
masking the identified heads moves it 0.000 (95\% CI \(\pm 0.063\)).

\textbf{Labeller invariance.} All results were re-scored under a second labeller that differs in its treatment of disclaimer-then-verdict constructions: the prescribed-set null holds in 4/4 model \(\times\) labeller combinations, with the sign of the change agreeing within each model, and the complement-set result is unchanged. Head selection is separately invariant --- the top-\(K\) set difference is identical in every cell. Both checks and the limits of the first are in Appendices~\ref{app:three-way-comparison} and~\ref{app:head-selection-invariance}.

\section{Discussion}\label{sec:discussion}\label{discussion}

\subsection{What the statistic measures}\label{sec:what-the-statistic-measures-and-what-c}\label{sec:two-circularities-one-of-them-in-the-e}

Rejecting the length hypothesis does not establish that the heads encode fairness. What remains is not attributable to the fairness construct, because within any single prompt the label, the response length and the presence of a protected-group token are collinear, the last partly by definition of the labelling rule. This is not a claim that bias is absent from attention heads, but a claim about what this procedure establishes on its own data: nothing that distinguishes the target construct from two surface correlates and one property of the prompt template. The failure is not repairable by better estimation: residualising length out of the contrast would remove the contrast. What would repair it is a different \textbf{elicitation} --- scoring log-probabilities over fixed candidates, stratifying the contrast by length, or holding the prompt constant across label conditions --- none of which is used in this family.

\textbf{Two circularities, one of them in the equation.} Heads are selected by a judge's fair/unfair label and the intervention is evaluated by the same label. This is \emph{construct} circularity rather than data leakage: it survives the original's disjoint selection and evaluation sets, because one instrument defines both target and test. The field identified this early --- \citet{yang2023biasahead} observe that their statistic is the gradient of SEAT while their evaluation is SEAT, and re-evaluate under an independent metric --- and their successors do not inherit the practice. A second
circularity sits inside the scoring equation: \(\bar v_{\text{ref}}\) is
constructed from the response tokens and the score is then summed over response
tokens, so every head is measured against a target it partly wrote. Both are compounded by the collinearity above: selection and evaluation partly measure the same surface properties, so a method can appear to work by shifting response form at either stage (Sections~\ref{sec:causal-evaluation}--\ref{sec:ablation-degraded-generation-is-scored}).

\subsection{What the complement set shows}\label{sec:what-the-complement-set-shows}\label{what-the-complement-set-shows}

The complement of the prescribed set --- heads ranking highly for refusal responses but not committal ones, under the identical top-\(K\) construction --- produces a clean unidirectional effect: zero counter-directional flips against 21 supporting ones across the stored arms under the primary labeller, \textbf{replicating across near-disjoint reselections} sharing 2--3 of 10 heads (\(p = 1.75\times10^{-5}\)). One counter-directional
flip appears in the entire study, in a reselected arm under the primary
labeller; we report it because it is what makes the surrounding zeros
meaningful.

This licenses the causal null: the same score, construction, masking code, evaluation set and labeller that yield no effect for the prescribed set yield a replicated effect for its complement, so the null is a property of that set rather than of our scale. It is the positive control \citet{kendiukhov2026systematicevaluationsinglecellfoundation} identifies as missing from ablation-based audits, and differs from the closest precedent: \citet{yang2023biasahead} mask the bottom of the same ranking under the same objective, testing whether the ranking is ordered, where ours is the top under the \emph{opposite} label, testing whether the pipeline detects anything at all.

\textbf{The asymmetry is not general, and its failure is informative.} Under one of the two reference readings on Qwen the prescribed set and its disjoint complement produce identical profiles; the asymmetry holds in two of three evaluable cells. We read this as favouring the interpretation that the contrast carries information about refusal behaviour rather than stereotype content: if both sides of the set difference select heads whose ablation destroys refusal, the statistic is not separating bias-carrying from fairness-carrying heads. We claim the control, not the mechanism.

\subsection{Scope, and what this family should report}\label{sec:the-family-has-seen-both-hazards}\label{sec:scope}\label{sec:what-a-method-in-this-family-should-re}

\textbf{The family has seen both hazards and acted on neither.} \citet{dasu2026attention} state that dropping attention heads increases perplexity and lowers measured bias because degenerate text is low-bias by default, their tables showing random pruning with the best bias score in its row at a perplexity six orders of magnitude above baseline, and \citet{zayed2023fairnessawarestructuredpruningtransformers} report random baselines producing visibly less coherent continuations on the bias-evaluation prompts. Both are qualitative remarks in appendices, and no paper in the family measures generation quality on its evaluation set, quantifies a degeneracy rate, or excludes broken generations from the denominator. On nulls, the random arms that exist are a point estimate at one pruning ratio or the best of several draws; our band moves refusal behaviour in both directions, so a single draw is uninformative about magnitude and sign alike.

\textbf{Scope.} Response length is a confound only where the statistic is computed over generated responses partitioned by a judge label; minimal-pair methods \citep{jahan2026localizingrepairingbiastransformer,yang2023biasahead} equalise response form by construction. The identifiability problem is broader: any procedure partitioning outputs by a judge's label inherits whatever that label is collinear with, and minimal-pair designs trade one collinearity for another --- with only the attribute word varying, the label is near-definitionally the attribute token's presence, the degeneracy we measure at \(\phi = 0.95\). We do not read our result as confined to the one design we measured directly.

\textbf{What a method in this family should report.} Three diagnostics, none currently standard: a control map built from the same per-example scores under an explicit alternative partition; a null distribution of ten or more matched component sets, reported as a band; and a degeneracy screen with an explicit denominator convention on the evaluation set itself --- plus, where the construction permits, a complement-set positive control (Section~\ref{sec:causal-evaluation}). We add a fourth, which our own results establish and which we did not anticipate. \textbf{A correlational control map is necessary and not sufficient.} Ours returned \(\rho = -0.950\) for an alternative hypothesis that a subsequent intervention rejected, and every additional correlational control we ran --- across two models, three persona strengths, two style restrictions and a fixed scoring window --- agreed with the wrong answer. Where a candidate confound is identified the coupling must be broken experimentally: a correlational diagnostic can establish that a statistic is confounded, never with what.

\section{Conclusion}\label{sec:conclusion}\label{conclusion}

We asked what a contrastive head-attribution statistic for social bias measures.
The answer we first arrived at was wrong, and the way it was wrong is the
paper's most transferable result: a control map built under an explicit
alternative hypothesis --- the diagnostic this literature lacks --- returned
\(-0.950\) with response length, survived every further correlational control we
could construct, and was rejected by the first intervention we ran. What the statistic tracks is prompt identity; what it cannot be shown to track is the fairness construct, which no method in this family runs the experiment to separate. The substantive claims fail independently. None of this shows that social bias cannot be localised in attention heads, only that this procedure does not establish that it has been --- and that the diagnostics which would reveal as much cost minutes (Section~\ref{sec:what-a-method-in-this-family-should-re}).

\section*{Limitations}\label{limitations}

\textbf{Scale and scope.} Our identification set is 60 questions against
the original's 2,516, and we use a validated rule-based labeller rather
than an LLM judge. We cannot fully separate ``the method does not
replicate'' from ``the reduction loses the effect''; map reliability
above 0.95 and the complement-set positive control (Section~\ref{sec:causal-evaluation}) argue
against the latter, but the objection cannot be eliminated. The length
hypothesis is tested on one member of the generative, judge-labelled
subfamily --- the only one we identified --- so the construct-validity
critique generalises across the family while the intervention does not.

\textbf{Two models}, both instruction-tuned, both 7--8B, both 4-bit NF4
with \(W_O\) dequantised for scoring. Quantisation effects on head-level
attribution were not assessed.

\textbf{The dependent variable.} Agreement with human judgment on a
blind stratified sample of the evaluation data is 0.867 {[}0.794,
0.916{]}, with a false-unfair rate of 0.120 (Section~\ref{sec:response-labelling-and-validation}). Annotation is
by a single author; no inter-annotator agreement is available. The
labeller reads a fixed 250-character window, and because masking more
than doubles response length, the prescribed arm exceeds that window
three times as often as baseline. Correcting for the resulting
misclassification increases the reported $\Delta$ under every error model
examined (Section~\ref{sec:causal-evaluation}), so the reported value is a lower bound --- but
that correction rests on sensitivity and specificity estimated from
eight observations each.

\textbf{Limits of our own identifiability analysis.} The collinearity
between the fairness label and protected-group-token presence
(\(\phi = 0.95\)) is measured in the terse condition alone (\(n = 60\)),
where a ten-token refusal has little room to name both groups; the
entailment holds everywhere by construction of the labelling rule, but
its converse is arm-dependent and unmeasured elsewhere. The prompt-identity result conflates prompt wording, instruction position and response format, which vary together across the four conditions: prompt identity carries the variance, but which property does so is not identified. CoT token counts are censored at the generation cap, so that arm contributes labels but not continuous length.

\textbf{Specification and implementation.} The source underspecifies the
reference vector, states the averaging set inconsistently, gives an
equation for the head set that contradicts its own figure caption, and
does not state \(k\) (Appendix~\ref{app:deviations-from-the-source-specifi}). We implement both reference
readings and report both; every conclusion holds under either, though
the \emph{descriptive} characterisation of the layer distribution does
not --- under one reading Qwen's selectivity is significantly earlier
than chance, under the other it is indistinguishable from uniform. Our own averaging set is capped at 64 response tokens where the source specifies all of them, which is immaterial for DA responses and truncates CoT ones. Neither reading is confirmed against the authors' implementation, so we cannot exclude that a third recovers a different ranking.

\textbf{Measurement exposure.} The prescribed arm exceeds the labeller's 250-character window far more often than baseline (0.475 against 0.138 on Qwen) and agreement is lower beyond it. Correcting for the resulting misclassification increases \(\Delta\) on both models under every error model we examined, so the reported value is a lower bound, and the corrected \(\Delta\) remains inside the random null band in all eight model \(\times\) error-model combinations; that correction rests on sensitivity and specificity estimated from eight observations each. The complement set is the \emph{least} window-exposed arm in the study.

\textbf{Verification not performed.} The scoring implementation was
compared against the source equation and its structural properties
recovered from the stored score arrays, but it was never independently
re-implemented from the equation text by a party without access to the
code. The ten-draw null band is layer-matched to the stored head sets
and does not transfer to the reselected sets, whose layer profiles
differ; percentile positions are therefore reported only where matched.

\textbf{CoT is not measurable with this dependent variable.} 20 of 60 and 24 of 60 CoT responses were unclassifiable on the two models, so no CoT quantity in this paper rests on a full evaluation set. The elicitation-dependence of head attribution (Appendix~\ref{app:elicitation-dependence-of-head-attri}) inherits that limit: it rests on nine unfair examples selected by opening-classifiability, and is not computable on Llama at \(n = 6\).

\textbf{The degeneracy screen is conservative}, missing typo-level
corruption, and two of its five signals produce false positives on long
structured text.

\section*{Ethical considerations}

\textbf{Content.} Our elicitation deliberately prompts models to make
stereotyped comparisons between social groups, and a majority of the responses
we collect commit to one group on a trait. We treat those responses only as
measurement data. We report no generated text that names a real individual, and
the question set is purpose-built and contains no personal data.

\textbf{What this paper does and does not argue.} Our findings concern the
validity of one attribution procedure. They are not evidence that social bias
is absent from these models, that it cannot be localised in attention heads, or
that mitigation is unnecessary. The behavioural premise the paradigm rests on
reproduces on both models
(Section~\ref{sec:the-behavioural-premise-replicates}); what does not reproduce
is the mechanistic account of it. A reader who took this paper as licence to
stop measuring bias would be reading it backwards.

\textbf{Risk of the method we audit.} The hazard we document in
Section~\ref{sec:ablation-degraded-generation-is-scored} has a deployment
consequence: an ablation that degrades generation can register as a reduction
in measured unfairness. A practitioner applying such a method and reading the
metric alone could ship a more damaged model believing it to be a fairer one.
This is the main practical reason we recommend a generation-quality screen with
an explicit denominator convention.

\textbf{Models and licensing.} Qwen-2.5-7B-Instruct \citep{qwen2025} and
Llama-3.1-8B-Instruct \citep{llama3} were used for research purposes only,
consistent with the licences under which they are released. \emph{Confirm the
exact licence names from each model's model card before the camera-ready.}

\textbf{Annotation.} All human labelling reported here was performed by an
author. No external annotators were recruited, no participant data was
collected, and no institutional review was sought or required.

\bibliographystyle{acl_natbib}
\bibliography{references}

\begin{thebibliography}{34}
\providecommand{\natexlab}[1]{#1}

\bibitem[{Adebayo et~al.(2018)Adebayo, Gilmer, Muelly, Goodfellow, Hardt, and Kim}]{adebayo2018sanity}
Julius Adebayo, Justin Gilmer, Michael Muelly, Ian Goodfellow, Moritz Hardt, and Been Kim. 2018.
\newblock Sanity checks for saliency maps.
\newblock \emph{Advances in Neural Information Processing Systems}, 31.

\bibitem[{Adebayo et~al.(2022)Adebayo, Muelly, Abelson, and Kim}]{adebayo2022post}
Julius Adebayo, Michael Muelly, Harold Abelson, and Been Kim. 2022.
\newblock Post hoc explanations may be ineffective for detecting unknown spurious correlation.
\newblock In \emph{International Conference on Learning Representations}.

\bibitem[{Adi et~al.(2016)Adi, Kermany, Belinkov, Lavi, and Goldberg}]{adi2016fine}
Yossi Adi, Einat Kermany, Yonatan Belinkov, Ofer Lavi, and Yoav Goldberg. 2016.
\newblock Fine-grained analysis of sentence embeddings using auxiliary prediction tasks.
\newblock \emph{arXiv preprint arXiv:1608.04207}.
\newblock ICLR 2017.

\bibitem[{Arditi et~al.(2024)Arditi, Obeso, Syed, Paleka, Panickssery, Gurnee, and Nanda}]{arditi2024refusal}
Andy Arditi, Oscar Obeso, Aaquib Syed, Daniel Paleka, Nina Panickssery, Wes Gurnee, and Neel Nanda. 2024.
\newblock Refusal in language models is mediated by a single direction.
\newblock \emph{Advances in Neural Information Processing Systems}, 37:136037--136083.

\bibitem[{Blodgett et~al.(2020)Blodgett, Barocas, Daum{\'e}~III, and Wallach}]{blodgett2020}
Su~Lin Blodgett, Solon Barocas, Hal Daum{\'e}~III, and Hanna Wallach. 2020.
\newblock Language (technology) is power: A critical survey of ``bias'' in {NLP}.
\newblock In \emph{ACL}.

\bibitem[{Blodgett et~al.(2021)Blodgett, Lopez, Olteanu, Sim, and Wallach}]{blodgett2021}
Su~Lin Blodgett, Gilsinia Lopez, Alexandra Olteanu, Robert Sim, and Hanna Wallach. 2021.
\newblock Stereotyping {N}orwegian salmon: An inventory of pitfalls in fairness benchmark datasets.
\newblock In \emph{ACL}.

\bibitem[{Cronbach and Meehl(1955)}]{cronbach1955}
Lee~J. Cronbach and Paul~E. Meehl. 1955.
\newblock Construct validity in psychological tests.
\newblock \emph{Psychological Bulletin}, 52(4):281--302.

\bibitem[{Dahlem et~al.(2026)Dahlem, Maniloff, and Misiura}]{dahlem2026selfattentiontransportlimitssymmetric}
Dominik Dahlem, Diego Maniloff, and Mac Misiura. 2026.
\newblock \href {https://arxiv.org/abs/2605.04893} {Self-attention as transport: Limits of symmetric spectral diagnostics}.
\newblock \emph{Preprint}, arXiv:2605.04893.

\bibitem[{Dasu et~al.(2026)Dasu, Rashid, Gupta, Tizpaz-Niari, and Tan}]{dasu2026attention}
Vishnu~Asutosh Dasu, Md~Rafi~Ur Rashid, Vipul Gupta, Saeid Tizpaz-Niari, and Gang Tan. 2026.
\newblock Attention pruning: Automated fairness repair of language models via surrogate simulated annealing.
\newblock In \emph{Proceedings of the 2026 IEEE/ACM 48th International Conference on Software Engineering}, pages 2824--2836.

\bibitem[{Dettmers et~al.(2023)Dettmers, Pagnoni, Holtzman, and Zettlemoyer}]{dettmers2023qlora}
Tim Dettmers, Artidoro Pagnoni, Ari Holtzman, and Luke Zettlemoyer. 2023.
\newblock \href {https://arxiv.org/abs/2305.14314} {{QL}o{RA}: Efficient finetuning of quantized {LLM}s}.
\newblock \emph{Preprint}, arXiv:2305.14314.

\bibitem[{Grattafiori et~al.(2024)Grattafiori, Dubey, Jauhri et~al.}]{llama3}
Aaron Grattafiori, Abhimanyu Dubey, Abhinav Jauhri, et~al. 2024.
\newblock \href {https://arxiv.org/abs/2407.21783} {The llama 3 herd of models}.
\newblock \emph{Preprint}, arXiv:2407.21783.

\bibitem[{Han et~al.(2026)Han, Song, Ding, Li, Fang, Li, Liu, Chen, and Wang}]{han2026debiasing}
Tingxu Han, Wei Song, Ziqi Ding, Ziming Li, Chunrong Fang, Yuekang Li, Dongfang Liu, Zhenyu Chen, and Zhenting Wang. 2026.
\newblock Debiasing {LLM}s by masking unfairness-driving attention heads.
\newblock In \emph{Findings of the Association for Computational Linguistics: ACL 2026}, pages 37980--37992.

\bibitem[{Hewitt and Liang(2019)}]{hewitt2019designing}
John Hewitt and Percy Liang. 2019.
\newblock Designing and interpreting probes with control tasks.
\newblock In \emph{EMNLP-IJCNLP}, pages 2733--2743.

\bibitem[{Huang et~al.(2025)Huang, Zhang, Yue, Li, Zhang, and Liu}]{rdsha2025}
Chao Huang, Zefeng Zhang, Juewei Yue, Quangang Li, Chuang Zhang, and Tingwen Liu. 2025.
\newblock \href {https://arxiv.org/abs/2508.19697} {Safety alignment should be made more than just a few attention heads}.
\newblock \emph{Preprint}, arXiv:2508.19697.

\bibitem[{Islam et~al.(2026)Islam, Borenstein, Pawar, Yu, Arora, and Augenstein}]{islam2026biasgymsimplegeneralizableframework}
Sekh~Mainul Islam, Nadav Borenstein, Siddhesh~Milind Pawar, Haeun Yu, Arnav Arora, and Isabelle Augenstein. 2026.
\newblock \href {https://arxiv.org/abs/2508.08855} {{BiasGym}: A simple and generalizable framework for analyzing and removing biases through injection}.
\newblock \emph{Preprint}, arXiv:2508.08855.

\bibitem[{Jahan(2026)}]{jahan2026localizingrepairingbiastransformer}
Sigma Jahan. 2026.
\newblock \href {https://arxiv.org/abs/2607.12863} {Toward localizing and repairing bias in transformer attention heads}.
\newblock \emph{Preprint}, arXiv:2607.12863.
\newblock ICSME 2026, NIER track.

\bibitem[{Kendiukhov(2026)}]{kendiukhov2026systematicevaluationsinglecellfoundation}
Ihor Kendiukhov. 2026.
\newblock \href {https://arxiv.org/abs/2602.17532} {Systematic evaluation of single-cell foundation model interpretability reveals attention captures co-expression rather than unique regulatory signal}.
\newblock \emph{Preprint}, arXiv:2602.17532.

\bibitem[{Kumar et~al.(2022)Kumar, Tan, and Sharma}]{kumar2022probing}
Abhinav Kumar, Chenhao Tan, and Amit Sharma. 2022.
\newblock Probing classifiers are unreliable for concept removal and detection.
\newblock \emph{Advances in Neural Information Processing Systems}, 35:17994--18008.

\bibitem[{M{\'e}loux et~al.(2025{\natexlab{a}})M{\'e}loux, Dirupo, Portet, and Peyrard}]{meloux2025dead}
Maxime M{\'e}loux, Giada Dirupo, Fran{\c{c}}ois Portet, and Maxime Peyrard. 2025{\natexlab{a}}.
\newblock The dead salmons of {AI} interpretability.
\newblock \emph{arXiv preprint arXiv:2512.18792}.

\bibitem[{M{\'e}loux et~al.(2025{\natexlab{b}})M{\'e}loux, Maniu, Portet, and Peyrard}]{meloux2025everything}
Maxime M{\'e}loux, Silviu Maniu, Fran{\c{c}}ois Portet, and Maxime Peyrard. 2025{\natexlab{b}}.
\newblock Everything, everywhere, all at once: Is mechanistic interpretability identifiable?
\newblock In \emph{International Conference on Learning Representations}, pages 32900--32923.

\bibitem[{M{\'e}loux et~al.(2025{\natexlab{c}})M{\'e}loux, Peyrard, and Portet}]{meloux2025mechanistic}
Maxime M{\'e}loux, Maxime Peyrard, and Fran{\c{c}}ois Portet. 2025{\natexlab{c}}.
\newblock Mechanistic interpretability as statistical estimation: A variance analysis of {EAP-IG}.
\newblock \emph{arXiv preprint arXiv:2510.00845}.

\bibitem[{Quirke(2026)}]{quirke2026ablation}
Philip Quirke. 2026.
\newblock Necessary, decodable and reversible, yet not transferable: A stress test for attention-head role claims.
\newblock \emph{arXiv preprint arXiv:2606.08292}.

\bibitem[{{Qwen Team}(2025)}]{qwen2025}
{Qwen Team}. 2025.
\newblock \href {https://arxiv.org/abs/2412.15115} {Qwen2.5 technical report}.
\newblock \emph{Preprint}, arXiv:2412.15115.

\bibitem[{Shaikh et~al.(2023)Shaikh, Zhang, Held, Bernstein, and Yang}]{shaikh2023}
Omar Shaikh, Hongxin Zhang, William Held, Michael Bernstein, and Diyi Yang. 2023.
\newblock On second thought, let's not think step by step! bias and toxicity in zero-shot reasoning.
\newblock In \emph{ACL}.

\bibitem[{Tan et~al.(2024)Tan, Chanin, Lynch, Paige, Kanoulas, Garriga-Alonso, and Kirk}]{tan2024analysing}
Daniel Tan, David Chanin, Aengus Lynch, Brooks Paige, Dimitrios Kanoulas, Adri{\`a} Garriga-Alonso, and Robert Kirk. 2024.
\newblock Analysing the generalisation and reliability of steering vectors.
\newblock \emph{Advances in Neural Information Processing Systems}, 37:139179--139212.

\bibitem[{Tseng et~al.(2024)Tseng, Huang, Hsiao, Chen, Huang, Meng, and Chen}]{personasurvey2024}
Yu-Min Tseng, Yu-Chao Huang, Teng-Yun Hsiao, Wei-Lin Chen, Chao-Wei Huang, Yu~Meng, and Yun-Nung Chen. 2024.
\newblock Two tales of persona in {LLM}s: A survey of role-playing and personalization.
\newblock In \emph{Findings of the Association for Computational Linguistics: EMNLP 2024}, pages 16612--16631. Association for Computational Linguistics.

\bibitem[{Voita et~al.(2019)Voita, Talbot, Moiseev, Sennrich, and Titov}]{voita2019analyzing}
Elena Voita, David Talbot, Fedor Moiseev, Rico Sennrich, and Ivan Titov. 2019.
\newblock Analyzing multi-head self-attention: Specialized heads do the heavy lifting, the rest can be pruned.
\newblock In \emph{ACL}, pages 5797--5808.

\bibitem[{Wolf et~al.(2020)Wolf, Debut, Sanh, Chaumond, Delangue, Moi, Cistac, Rault, Louf, Funtowicz et~al.}]{wolf2020transformers}
Thomas Wolf, Lysandre Debut, Victor Sanh, Julien Chaumond, Clement Delangue, Anthony Moi, Pierric Cistac, Tim Rault, Remi Louf, Morgan Funtowicz, et~al. 2020.
\newblock Transformers: State-of-the-art natural language processing.
\newblock In \emph{Proceedings of the 2020 Conference on Empirical Methods in Natural Language Processing: System Demonstrations}, pages 38--45. Association for Computational Linguistics.

\bibitem[{Xu(2026)}]{xu2026closurevalidatedcircuitdiscoveryattention}
Yongzhong Xu. 2026.
\newblock \href {https://arxiv.org/abs/2606.09607} {Closure-validated circuit discovery in attention heads: Co-activation proposes, ablation disposes}.
\newblock \emph{Preprint}, arXiv:2606.09607.

\bibitem[{Yang et~al.(2023)Yang, Duan, Abbasi, Lalor, and Tam}]{yang2023biasahead}
Yi~Yang, Hanyu Duan, Ahmed Abbasi, John~P. Lalor, and Kar~Yan Tam. 2023.
\newblock \href {https://arxiv.org/abs/2311.10395} {Bias a-head? analyzing bias in transformer-based language model attention heads}.
\newblock \emph{Preprint}, arXiv:2311.10395.

\bibitem[{Yin et~al.(2025)Yin, Leong, Yang, Huang, Li, Wang, Yoon, Xing, Yu, and Gu}]{refusalcliff2025}
Qingyu Yin, Chak~Tou Leong, Linyi Yang, Wenxuan Huang, Wenjie Li, Xiting Wang, Jaehong Yoon, Yun Xing, Xing Yu, and Jinjin Gu. 2025.
\newblock \href {https://arxiv.org/abs/2510.06036} {Refusal falls off a cliff: How safety alignment fails in reasoning?}
\newblock \emph{Preprint}, arXiv:2510.06036.

\bibitem[{Zayed et~al.(2023)Zayed, Mordido, Shabanian, Baldini, and Chandar}]{zayed2023fairnessawarestructuredpruningtransformers}
Abdelrahman Zayed, Goncalo Mordido, Samira Shabanian, Ioana Baldini, and Sarath Chandar. 2023.
\newblock \href {https://arxiv.org/abs/2312.15398} {Fairness-aware structured pruning in transformers}.
\newblock \emph{Preprint}, arXiv:2312.15398.
\newblock AAAI 2024.

\bibitem[{Zhao et~al.(2025)Zhao, Fu, Schaeffer, Sharma, and Barez}]{zhao2025chain}
Jianli Zhao, Tingchen Fu, Rylan Schaeffer, Mrinank Sharma, and Fazl Barez. 2025.
\newblock Chain-of-thought hijacking.
\newblock \emph{arXiv preprint arXiv:2510.26418}.

\bibitem[{Zhou et~al.(2024)Zhou, Feng, Zhu, Qian, and Mao}]{zhou2024unibiasunveilingmitigatingllm}
Hanzhang Zhou, Zijian Feng, Zixiao Zhu, Junlang Qian, and Kezhi Mao. 2024.
\newblock \href {https://arxiv.org/abs/2405.20612} {{UniBias}: Unveiling and mitigating {LLM} bias through internal attention and {FFN} manipulation}.
\newblock \emph{Preprint}, arXiv:2405.20612.
\newblock NeurIPS 2024.

\end{thebibliography}

\appendix

\section{Additional evaluation sets and full degradation analysis}\label{app:additional-evaluation-sets-and-ful}\label{appendix-a-additional-evaluation-sets-and-full-degradation-analysis}

\subsection{Evaluation sets}\label{app:evaluation-sets}\label{a.1-evaluation-sets}

Three evaluation sets were used across the study. The main text reports the
\textbf{balanced set} (80 questions; 8 protected-attribute pairs $\times$ 10 traits, no
trait--pair overlap with identification). Two earlier sets are reported here.

\textbf{EV-1 (36 questions, mixed).} The held-out remainder of the identification
grid. Underpowered and compositionally uncontrolled; used only in early
piloting. Baseline 28/36 = 0.778, prescribed-set 25/36 = 0.694
(\(\Delta = -0.083\), n.s.).

\textbf{EV-2 (100 questions, skewed).} Drawn from a wider trait grid but
over-weighting softer comparisons (61 of 100 involve protected attributes; only
one race-related item). Its high baseline (0.980) leaves almost no refusal
headroom, which is precisely what makes it informative for the degradation
analysis below: with only two baseline refusals available to destroy, any
apparent \emph{debiasing} movement on this set is suspect by construction.

The prescribed-set null replicates on all three sets, at baselines spanning 0.708--0.980: \(\Delta = -0.083\), \(+0.000\), \(+0.013/+0.086\) (EV-1, EV-2, balanced Qwen/Llama), none significant. \emph{The EV-1 and EV-2 figures predate the labeller correction and are reported as scored at the time; the balanced-set values are under the primary labeller used throughout.}

\subsection{Full degradation analysis (EV-2, Qwen)}\label{app:full-degradation-analysis-ev-2-qwe}\label{a.2-full-degradation-analysis-ev-2-qwen}

Six arms $\times$ 100 questions. Rate is unfair/(unfair+fair); unclassifiable counts
reported separately.

\begin{table*}[t]
\centering
\small
\begin{tabular}{@{}lllllll@{}}
\toprule
arm & unfair & excl. & rate & flips u$\rightarrow$f & degenerate & mean tok \\
\midrule
baseline & 98 & 0 & 0.980 & --- & 0 & 19.2 \\
prescribed (differ-unfair) & 98 & 0 & 0.980 & 0 & 0 & 44.8 \\
reverse set & 99 & 0 & 0.990 & 0 & 0 & 18.4 \\
random-0 & 80 & 18 & 0.976 & 19 & \textbf{34} & 30.2 \\
random-1 & 97 & 1 & 0.980 & 1 & 0 & 30.8 \\
random-2 & 94 & 5 & 0.989 & 5 & \textbf{17} & 53.6 \\
\bottomrule
\end{tabular}
\caption{Full degradation analysis on EV-2 (Qwen), six arms $\times$ 100 questions. Rate is unfair/(unfair+fair); unclassifiable counts are reported separately.}
\end{table*}

Omnibus across six arms: \(\chi^2 = 0.94\), \(p = 0.967\). The prescribed set is
item-identical to baseline: 100/100 verdicts unchanged while mean length rises
from 19.2 to 44.8 tokens (\(p < 10^{-4}\)). Verbosity is not specific to the
prescribed set --- random-2 lengthens more (+34.4 vs +25.6).

\textbf{Flips are breakage.} Random-0's 19 flips: 14/19 (74\%) degenerate vs 20/81
(25\%) among non-flips (Fisher \(p = 0.0001\)). Random-2's 5 flips: 4/5 (80\%) vs
13/95 (14\%) (\(p = 0.0027\)). Clean-only reanalysis (all degenerate responses
dropped): \(\chi^2 = 0.59\), \(p = 0.9886\) --- the only significant behavioural
effect on this set vanishes entirely.

\textbf{Failure taxonomy} (five-signal screen; conservative, so rates are lower
bounds): doubled word (``People from the the north\ldots{}''), line repeated $\geq$3$\times$,
lexical diversity \textless{} 0.55 at $\geq$20 tokens, verbatim prompt echo, 4-gram repeated
$\geq$3$\times$. Random-0's failures are dominated by doubled words (34/34); random-2
spans all five signals including full collapse into a repeated token. On long
structured text, \texttt{low\_diversity} and \texttt{ngram\_loop} false-positive on legitimate
formatting (18/61 and 18/61 on identification CoT vs 3 \texttt{dup\_word}), so for
long-form text only \texttt{dup\_word} and \texttt{line\_loop} are trustworthy.

\subsection{Ten-draw random null (balanced set)}\label{app:ten-draw-random-null-balanced-set}\label{a.3-ten-draw-random-null-balanced-set}

\textbf{the table above.} Per-draw refusal counts and behavioural flips, Qwen (baseline:
19 refusals). Arms exceeding 20\% unclassifiable are marked broken and excluded
from the null band.

\begin{table*}[t]
\centering
\small
\begin{tabular}{@{}llllll@{}}
\toprule
draw & refusals & u$\rightarrow$f & f$\rightarrow$u & excl. & McNemar (Holm) \\
\midrule
0 & 9 & 0 & 7 & 3 & 0.0781 \\
1 & 15 & 1 & 2 & 5 & 1.0000 \\
2 & 0 & 0 & 19 & 1 & \textbf{0.0000} \\
3 & 27 & 9 & 0 & 1 & \textbf{0.0312} \\
4 & 5 & 0 & 12 & 3 & \textbf{0.0044} \\
5 & 12 & 10 & 1 & \textbf{54 (broken)} & --- \\
6 & 2 & 1 & 16 & 8 & \textbf{0.0027} \\
7 & 12 & 1 & 8 & 1 & 0.1562 \\
8 & 27 & 8 & 0 & 2 & \textbf{0.0469} \\
9 & 20 & 4 & 1 & 2 & 1.0000 \\
\bottomrule
\end{tabular}
\caption{Ten-draw random null on the balanced set: per-draw refusal counts and behavioural flips on Qwen (baseline 19 refusals). Arms exceeding 20\% unclassifiable are marked broken and excluded from the band. Scored under the predecessor labeller.}
\end{table*}

Six clean draws destroyed refusals, three created them; five survive Holm
correction --- in \textbf{both} directions (draws 2, 4, 6 destroy; 3 and 8 create).
\emph{This table is scored under the predecessor labeller.} Clean-draw rate band: 0.654--1.000, mean 0.833, sd 0.121. Under the primary labeller used throughout the main text the band is 0.641--1.000 (Section~\ref{sec:causal-evaluation}); the broken-arm set and the five Holm-surviving draws, in both directions, are identical under the two labellers, so only the band, mean and standard deviation differ. Llama (not shown):
band 0.631--0.884 over eight clean draws, two broken at 27.5\% and 28.7\%
unclassifiable.

\subsection{Labeller definitions and validation detail}\label{app:labeller-definitions-and-validatio}\label{a.4-labeller-definitions-and-validation-detail}

Primary labeller (\texttt{label4}): reads the opening 250 characters; \emph{unfair} if a
group name opens the response or exactly one group is named without hedging;
\emph{fair} on a refusal/hedging marker; unclassifiable otherwise. Validation:
Qwen 100 responses (95/100 pre-adjustment, 0 directional errors; 100/100
post-adjustment, optimistic as fitted to those rows; short/long-half agreement
1.000/1.000; 72/72 agreement with an independent order-based rule). Llama:
30 held-out DA responses sampled after finalisation with automated labels
hidden --- 0.900, 0 false-unfair, 0 false-fair, 3 unclassifiable; short/long
1.000/0.800.

Secondary labeller (\texttt{label6}): reversal-aware variant (refusal marker followed
by a reversal cue and a group name $\rightarrow$ unfair), with an extended refusal
vocabulary. Development-set agreement 0.962; held-out 0.900 with one
false-fair. Rejected as primary; used for the invariance check (the labeller-invariance section).

\begin{center}\rule{0.5\linewidth}{0.5pt}\end{center}

\section{Labeller development, failure audit, and reproducibility notes}\label{app:labeller-development-failure-audit}\label{appendix-b-labeller-development-failure-audit-and-reproducibility-notes}

\subsection{Development history}\label{app:development-history}\label{b.1-development-history}

\begin{table*}[t]
\centering
\footnotesize
\begin{tabular}{@{}>{\raggedright\arraybackslash}p{(\textwidth - 10\tabcolsep) * \real{0.2000}}>{\raggedright\arraybackslash}p{(\textwidth - 10\tabcolsep) * \real{0.2000}}>{\raggedright\arraybackslash}p{(\textwidth - 10\tabcolsep) * \real{0.2000}}>{\raggedright\arraybackslash}p{(\textwidth - 10\tabcolsep) * \real{0.2000}}>{\raggedright\arraybackslash}p{(\textwidth - 10\tabcolsep) * \real{0.2000}}@{}}
\toprule
sheet & \(n\) & blind & source file & what it fed \\
\midrule
Qwen development & 100 & \textbf{no} & an earlier 4-arm evaluation file (144 responses) & initial agreement figure \\
Llama development & 50 & \textbf{no} & the Llama identification set & reversal-aware variant development \\
Llama held-out & 30 & yes & the Llama identification set, Direct-Answer only & variant rejection \\
\bottomrule
\end{tabular}
\caption{Three annotation sheets preceded the validation reported in Table~\ref{tab:labeller}. None was drawn from the files that produce the results below, and two displayed the rule's output to the annotator..}\label{tab:auto10}
\end{table*}

Consequences we record rather than defend. The Qwen sheet's
post-adjustment agreement was fitted to the rows it was measured on, and
the notebook's own comment describes it as optimistic. Neither model's
reported evaluation file was covered: the Qwen sheet came from a
four-arm file containing one random-ablation draw against the reported
file's ten, and both Llama sheets came from the identification set, so
the labeller was never checked on ablated Llama output before being used
to score ten ablation arms. Table~\ref{tab:labeller}'s blind stratified sample closes
this.

\subsection{Confusion matrix}\label{app:confusion-matrix}

Primary labeller (rows) against human judgment (columns), on the 120-response
blind validation sample of Table~\ref{tab:labeller}.

\begin{center}
\small
\begin{tabular}{@{}llll@{}}
\toprule
& unfair & fair & unclassifiable \\
\midrule
\textbf{unfair} & 83 & 3 & 0 \\
\textbf{fair} & 4 & 18 & 1 \\
\textbf{unclassifiable} & 4 & 4 & 3 \\
\bottomrule
\end{tabular}
\end{center}

\subsection{The failure audit}\label{app:the-failure-audit}\label{b.2-the-failure-audit}

All 3,932 stored responses across 11 files were scanned for three
failure modes.

\textbf{A --- terse negated refusals.} Constructions such as
\emph{``Leadership quality isn't defined by race''} contain no trigger
term and are too short to contain one, so they returned unclassifiable.
26 instances.

\textbf{B --- affirmative comparatives.} \emph{``I believe X are lazier
than Y''} names both groups, failing the opening match and then the
single-group rule, returning unclassifiable. 18 instances. Negated
comparatives (\emph{``not more X than''}) are refusals and must not be
caught; an initial patch without negation handling would have
mislabelled them, and was rejected before use.

\textbf{C --- the opening rule firing under negation.} \emph{``Rich
people are not inherently more intelligent than poor people\ldots{}''}
opens with a group name and was scored unfair. 15 instances. This is the
only pattern that moves responses \emph{between} fair and unfair rather
than out of the unclassifiable bucket, so it shifts rates directly, and
it is a false-unfair --- the error class the predecessor labeller's
validation reported as zero.

The three patterns push the length--label coupling in opposite
directions: A adds short-fair examples, C moves long responses from
unfair to fair. Their net effect on every reported correlation is at
most 0.004 (Section~\ref{sec:the-map-correlates-with-a-response-len}).

\textbf{Composition.} The rules are applied \emph{after} the degeneracy
screen. A rescue applied first would let corrupted text be scored fair
--- the hazard Section~\ref{sec:ablation-degraded-generation-is-scored} documents. One flagged response
(\emph{``Negotiation skills aren't not about race. They come from
experience, practice, and understanding of the the situation''}) is
degenerate and is correctly not rescued.

\subsection{Pattern precision}\label{app:pattern-precision}\label{b.3-pattern-precision}

Measured against a targeted sample of 56 flagged responses,
hand-labelled blind.

\begin{center}
\small
\begin{tabular}{@{}llll@{}}
\toprule
pattern & fired & correct & precision \\
\midrule
A & 26 & 21 & 0.808 \\
B & 12 & 10 & 0.833 \\
C & 7 & 5 & 0.714 \\
\bottomrule
\end{tabular}
\end{center}

Net across all 175 hand-labelled rows: 36 corrections, 4 regressions.

We report these rather than tuning against them. Revising the rules to
fit this sample would reproduce the defect the audit was conducted to
find. The confirmatory figure is Table~\ref{tab:labeller}'s blind
stratified agreement,
not per-pattern precision, and at these \(n\) the intervals do not
establish precision above 0.90 in any case.

One pattern-B false positive appears in the identification set: a
641-character chain-of-thought hedge containing a comparative,
blind-labelled unclassifiable. It is the sole source of the one-instance
discrepancy in the chain-of-thought ladder of Appendix~\ref{app:further-correlational-controls}.

\subsection{Three-way comparison}\label{app:three-way-comparison}\label{b.4-three-way-comparison}

\begin{table*}[t]
\centering
\footnotesize
\begin{tabular}{@{}>{\raggedright\arraybackslash}p{(\textwidth - 4\tabcolsep) * \real{0.5000}}>{\raggedright\arraybackslash}p{(\textwidth - 4\tabcolsep) * \real{0.5000}}@{}}
\toprule
finding & movement \\
\midrule
F1 map correlations & \(\Delta \leq 0.004\) on every row; split-half reliability identical \\
Layer distribution & centre-of-mass unchanged to 0.02 layers \\
Prescribed-set effect & $\Delta$ shrinks: Qwen +0.025 $\rightarrow$ +0.013; Llama +0.106 $\rightarrow$ +0.086; both remain non-significant \\
Complement-set effect & Qwen 0/9 $\rightarrow$ 0/8; Llama 0/13 unchanged; zero counter-directional throughout \\
Null bands & Qwen 0.654--1.000 $\rightarrow$ 0.641--1.000; Llama 0.631--0.884 $\rightarrow$ 0.603--0.886 \\
Broken-arm set & identical under all three \\
\bottomrule
\end{tabular}
\caption{Every reported number was recomputed under the predecessor labeller, the reversal-aware variant, and the primary labeller.}\label{tab:auto12}
\end{table*}

The two labellers share a refusal vocabulary and an opening-match rule, so they share the blind spots those induce: the check establishes robustness to one design choice, not labeller independence.

The reversal-aware variant was built to catch a disclaimer-then-verdict
pattern in Llama output. It scored 0.962 on its development sample and
0.900 on held-out data --- no better than its predecessor, with one
additional false-fair --- and was rejected as primary. It is retained as
the robustness column reported in Section~\ref{sec:the-outcome-measure-is-not-saturated}. It is not an independent instrument:
it shares a refusal vocabulary and an opening-match rule with the
predecessor and inherits both blind spots.

\subsection{Head selection invariance}\label{app:head-selection-invariance}\label{b.5-head-selection-invariance}

The labeller correction did not change which heads were selected.
Recomputing the top-\(K\) set difference under both labellers, on both
models, under both reference readings gives \textbf{identical head sets
in every cell}, with one head's difference in a single complement set.
On Qwen the rescued responses are unscored, so the partition is
unchanged by construction; on Llama the single reclassification shifts
the aggregate means too little to reorder the top 20.

\subsection{Selection provenance}\label{app:selection-provenance}\label{b.6-selection-provenance}

The stored Qwen head set was selected from aggregate score maps under a
predecessor labeller and the final-layer reference, where Section~\ref{sec:contribution-score}
designates the layer-matched reference primary. The stored Llama set
used the layer-matched reference and the reported labeller.
Reconstructing the Qwen selection under the described pipeline recovers
3 of 10 heads; correcting only the reference recovers 7 of 10; the
aggregate maps reproduce it exactly. The same reference discrepancy
accounts for the mixed conventions in the layer-distribution analysis (Section~\ref{sec:layer-wise-localisation-does-not-repro}). We report the causal evaluation under both references on both models (Section~\ref{sec:causal-evaluation}) rather than relying on either selection.

\subsection{Reproducibility notes}\label{app:reproducibility-notes}\label{b.7-reproducibility-notes}

Recorded because each would otherwise be found by a reader auditing the
code, and because the paper's argument concerns exactly this class of
unstated choice.

\begin{enumerate}
\def\labelenumi{\arabic{enumi}.}
\tightlist
\item
  \textbf{Masking-hook failures, twice.} An early experiment was voided
  when hooks were silently detached by an intervening analysis step. A
  later run was voided when the intervention-fidelity check itself
  cleared the mask before generation: the check function manages mask
  state, so passing it immediately before a loop does not establish the
  mask is live during it. Both runs were discarded and regenerated.
  Aggregate statistics from such a run are indistinguishable from a
  genuine null, which is why every masking arm now asserts the mask is
  armed at the start and re-asserts periodically.
\item
  \textbf{A truncation that reads as a bug and is not.} The complement
  set is truncated by \texttt{sorted(...)} on the
  \texttt{(layer,\ head)} tuple rather than by score rank. Because the
  two top-\(K\) sets have equal cardinality, their set differences do
  too, so the slice is a no-op and never removes a head.
\item
  \textbf{The primary labeller is defined twice} in the Qwen notebook;
  the copies are byte-equivalent modulo whitespace, with identical
  refusal vocabulary.
\item
  \textbf{An independence check that was not independent.} A verification pass intended to be independent retained state referencing this project, so it is not counted as independent verification anywhere in this paper.
\item
  \textbf{Score arrays are stored per example}, shape
  \([n_{\text{ex}}, L, H]\), rather than accumulated. Every reliability,
  placebo, bootstrap and re-split analysis in this paper is arithmetic
  on those arrays and required no additional model evaluation.
\end{enumerate}

\subsection{Deviations from the source specification}\label{app:deviations-from-the-source-specifi}\label{b.8-deviations-from-the-source-specification}

\begin{table*}[t]
\centering
\footnotesize
\begin{tabular}{@{}>{\raggedright\arraybackslash}p{(\textwidth - 6\tabcolsep) * \real{0.3333}}>{\raggedright\arraybackslash}p{(\textwidth - 6\tabcolsep) * \real{0.3333}}>{\raggedright\arraybackslash}p{(\textwidth - 6\tabcolsep) * \real{0.3333}}@{}}
\toprule
item & source & ours \\
\midrule
averaging set \(R\) & all response tokens & first 64 \\
\(\bar v_{\text{ref}}\) & undefined; ``first few tokens'' contradicts the stated mean over \(R\) & first 8 response positions, unit-normalised, two layer readings \\
\(z_{r,h}\) & ``the value vector of head \((l,h)\)'' & the per-head slice of the attention output \\
head set & Eq. 4 states top-\(k\) of one map; the text and figure describe a set difference & set difference \\
\(k\) & unspecified & \(K = 20\) \\
\bottomrule
\end{tabular}
\caption{Established by comparison against the source paper's Equation 2 rather than against our own description.}\label{tab:auto13}
\end{table*}

On \(z_{r,h}\): the quantity \(W_{O,h}\) left-multiplies is the
attention output, not the value vector, and only that reading makes the
score a contribution to the residual stream. Grouped-query attention
settles it --- Qwen-2.5-7B has 28 query heads and 4 key--value heads per
layer, so the literal reading would yield 4 distinct scores per layer
rather than 28.

\subsection{Elicitation-dependence of head attribution}\label{app:elicitation-dependence-of-head-attri}

Head maps estimated under Direct-Answer and under Chain-of-Thought elicitation
are only weakly related. On Qwen the cross-style correlation, disattenuated for
split-half reliability, is \(r = 0.286\)--\(0.489\) depending on the reference
reading. The estimate rests on the nine unfair Chain-of-Thought examples that
the labeller could classify from their opening window, so it is reported as an
indication of magnitude rather than a measurement with usable precision. The
same quantity is not computable on Llama, where only six unfair
Chain-of-Thought examples survive classification. Section~\ref{sec:elicitation}
states the elicitation protocol and the Limitations section states what this
restricts.

\subsection{Further correlational controls on the head map}\label{app:further-correlational-controls}

\begin{table*}[t]
\centering
\small
\begin{tabular}{@{}>{\raggedright\arraybackslash}p{(\textwidth - 14\tabcolsep) * \real{0.1429}}>{\raggedright\arraybackslash}p{(\textwidth - 14\tabcolsep) * \real{0.1429}}>{\raggedright\arraybackslash}p{(\textwidth - 14\tabcolsep) * \real{0.1429}}>{\raggedright\arraybackslash}p{(\textwidth - 14\tabcolsep) * \real{0.1429}}>{\raggedright\arraybackslash}p{(\textwidth - 14\tabcolsep) * \real{0.1429}}>{\raggedright\arraybackslash}p{(\textwidth - 14\tabcolsep) * \real{0.1429}}>{\raggedright\arraybackslash}p{(\textwidth - 14\tabcolsep) * \real{0.1429}}@{}}
\toprule
model & persona & style & cells & reliability & \(\rho\)(length, pooled) & top-10 $\in$ length-bottom-10 \\
\midrule
Qwen & full & pooled & 57u/43f & 0.950 & \textbf{-0.950} & 9/10 \\
Qwen & mild & pooled & 26u/78f & 0.978 & -0.828 & 7/10 \\
Qwen & full & DA only & 48u/12f & 0.911 & -0.725 & 6/10 \\
Qwen & mild & DA only & 26u/34f & 0.965 & \textbf{-0.970} & \textbf{10/10} \\
Llama & full & pooled & 48u/42f & 0.959 & \textbf{-0.768} & 8/10 \\
\bottomrule
\end{tabular}
\caption{Head-map correlations, layer-matched reference.}\label{tab:maps}
\end{table*}

\begin{table*}[t]
\centering
\small
\begin{tabular}{@{}>{\raggedright\arraybackslash}p{(\textwidth - 14\tabcolsep) * \real{0.1429}}>{\raggedright\arraybackslash}p{(\textwidth - 14\tabcolsep) * \real{0.1429}}>{\raggedright\arraybackslash}p{(\textwidth - 14\tabcolsep) * \real{0.1429}}>{\raggedright\arraybackslash}p{(\textwidth - 14\tabcolsep) * \real{0.1429}}>{\raggedright\arraybackslash}p{(\textwidth - 14\tabcolsep) * \real{0.1429}}>{\raggedright\arraybackslash}p{(\textwidth - 14\tabcolsep) * \real{0.1429}}>{\raggedright\arraybackslash}p{(\textwidth - 14\tabcolsep) * \real{0.1429}}@{}}
\toprule
persona & tokens, unfair & tokens, fair & gap & AUROC length $\rightarrow$ label & \(n\) unfair & LOO swing \\
\midrule
full & 34.5 & 104.8 & +70.4 & 0.878 & 57 & 0.010 \\
mild & 13.7 & 93.3 & +79.7 & 0.956 & 26 & 0.015 \\
none & 13.0 & 85.8 & +72.8 & 0.996 & \textbf{5} & 0.004 \\
Llama, full & --- & --- & --- & 0.836 & 48 & 0.017 \\
\bottomrule
\end{tabular}
\caption{Response length by label under three persona strengths (Qwen, identification set). LOO swing is the largest change in AUROC from removing a single unfair example. No confidence interval is reported for the neutral-persona rung: its AUROC lies at the boundary, where the standard estimator's variance collapses and produces an interval that understates uncertainty. The rung rests on five unfair responses and is reported for completeness, not as support.}\label{tab:length}
\end{table*}

Section~\ref{sec:the-map-correlates-with-a-response-len} reports the control maps, their reliability and the head-set overlap. The remaining controls, every one of which agreed with the length hypothesis that Section~\ref{sec:testing-the-length-hypothesis-an-inter} went on to reject, are given here.

\textbf{It is not the estimator.} Fixing the scoring window at 12 tokens for
every example, which removes any dependence of the score on response length,
leaves the correlation essentially unchanged: of the reduction from -0.959 to
-0.712, sample attrition accounts for -0.237 and the window itself for
\textbf{-0.011, or 4.5\%}.

\textbf{It is not the prompt-style axis.} The DA-only rows of
Table~\ref{tab:maps} hold style constant, and the mild-persona DA-only cell
(26 unfair / 34 fair, the most balanced contrast in this study) gives the
strongest correlation with complete head-set overlap. Within a single style the
fairness label and response length are nearly the same partition of examples
(AUROC 0.956), so this correlation and that AUROC are one fact measured twice
rather than two findings.

\textbf{The elicitation segregates the label cells by style.} Chain-of-Thought
unfair counts fall 9 $\rightarrow$ 0 $\rightarrow$ 1 across the three persona
conditions under the primary labeller (9 $\rightarrow$ 0 $\rightarrow$ 0 under
its predecessor; the single neutral-condition instance is a verified labeller
false positive, Appendix~\ref{app:pattern-precision}), while the Direct-Answer
share gap between the two label cells stays near-constant (56, 56, 50
percentage points): the persona changes how often the model commits, not how
the cells are segregated by style. Under a neutral prompt the procedure yields
\textbf{5 unfair responses of 60} --- too few to construct a contrast at all.
The identification procedure requires an adversarial framing to generate its
own signal.

\textbf{A mechanism is available, and it motivates a hypothesis.} On the
evaluation arms the point-biserial correlation between the unfair label and
response length is -0.452 (\(p = 1.3\times10^{-8}\)) and length alone
classifies at AUC 0.835, with unfair responses averaging 25.2 tokens against
49.0 for fair ones. Committing to a group is terse; declining is verbose. A
statistic that separates the two response populations would also separate short
from long responses, and any head whose activity covaries with response length
would be selected.

\subsection{Measurement exposure and the direction of its bias}\label{app:measurement-exposure}

Masking's effect on response length interacts with the labeller's fixed
250-character reading window. The prescribed arm exceeds that window far
more often than baseline --- 0.475 against 0.138 on Qwen, 0.438 against
0.175 on Llama, with non-overlapping intervals --- and agreement is
lower beyond the window (0.696) than within it (0.907).

Correcting for the resulting misclassification \textbf{increases} $\Delta$ on
both models under every error model we examined: Qwen from +0.013 to a
range of {[}+0.013, +0.087{]}, Llama from +0.086 to {[}+0.086,
+0.161{]}. The direction is invariant because out-of-window exposure
degrades sensitivity and specificity together and the prescribed arm is
the more exposed arm, so the reported $\Delta$ is a lower bound. This rests on
out-of-window sensitivity and specificity estimated from 8 and 8
observations respectively and assumed to transfer across arms. The
corrected $\Delta$ remains inside the random null band in all eight model $\times$
error-model combinations. The complement set is the \emph{least}
window-exposed arm in the study (0.062, 0.087), so no window-based objection reaches Section~\ref{sec:the-complement-set}.

\subsection{Degeneracy, the screen, and the denominator}\label{app:degeneracy-and-the-denominator}

\textbf{It generalises.} On an earlier unbalanced evaluation set
(Appendix~\ref{app:additional-evaluation-sets-and-ful}), one random draw produced 19 unfair$\rightarrow$fair flips, the only
significant apparent debiasing effect on that set. That arm contained 34
degenerate responses of 100 against zero in baseline; 74\% of its flips
were degenerate against 25\% of its non-flips (\(p = 0.0001\)), and a
second draw showed 80\% against 14\% (\(p = 0.0027\)). Dropping every
degenerate response removes the effect entirely (\(\chi^2 = 0.59\),
\(p = 0.9886\)). The failures are token repetitions --- \emph{``People
from the the north\ldots{}''}, \emph{``\ldots overrepresented in in
tech\ldots{}''} --- that a rule-based labeller scores as non-committal.

Validity destruction is heterogeneous across draws: unclassifiable
counts range from 1 to 54 of 80 on Qwen and 4 to 23 on Llama, with one
and two arms respectively exceeding the exclusion threshold.

Our screen uses five conservative signals --- a doubled word, a line
repeated three or more times, lexical diversity below 0.55 at 20 or more
tokens, verbatim prompt echo, and a 4-gram repeated three or more times
--- and does not detect typo-level corruption, so reported rates are
lower bounds. The denominator convention matters as much as the screen:
scoring unclassifiable responses as ``not unfair'' credits breakage as
improvement, and did so in an earlier version of our own analysis before
it was caught (Appendix~\ref{app:reproducibility-notes}).

\subsection{Layer distribution of bias selectivity}\label{app:layer-distribution}

Under the layer-matched reference, bias selectivity sits significantly \textbf{earlier} than chance on both models: centre-of-mass 11.72 against a uniform null of 13.5 on Qwen (\(p = 0.011\) early, \(p = 0.989\) against late) and 13.67 against 15.5 on Llama (\(p = 0.021\); \(p = 0.979\)). The final four layers carry 0.067 and 0.073 of selectivity against null shares of 0.143 and 0.125, under half of chance on both.

The \emph{early} concentration is not invariant to the reference reading --- it holds under both on Llama (\(p = 0.002\) final-layer) but only under the layer-matched reading on Qwen, where the final-layer one is indistinguishable from uniform (\(p = 0.428\)). What survives both readings on both models is the negative claim: of the eight model \(\times\) reference \(\times\) labeller combinations we computed, seven place selectivity earlier than chance and \textbf{none shows the late concentration the method reports}. With map reliability above 0.95 this is not estimation noise.

\begin{table*}[t]
\centering
\small
\begin{tabular}{@{}>{\raggedright\arraybackslash}p{(\textwidth - 14\tabcolsep) * \real{0.1429}}>{\raggedright\arraybackslash}p{(\textwidth - 14\tabcolsep) * \real{0.1429}}>{\raggedright\arraybackslash}p{(\textwidth - 14\tabcolsep) * \real{0.1429}}>{\raggedright\arraybackslash}p{(\textwidth - 14\tabcolsep) * \real{0.1429}}>{\raggedright\arraybackslash}p{(\textwidth - 14\tabcolsep) * \real{0.1429}}>{\raggedright\arraybackslash}p{(\textwidth - 14\tabcolsep) * \real{0.1429}}>{\raggedright\arraybackslash}p{(\textwidth - 14\tabcolsep) * \real{0.1429}}@{}}
\toprule
model & \(\bar v_{\text{ref}}\) & CoM layer & uniform null & \(p\)(late) & \(p\)(early) & final-4 share (null) \\
\midrule
Qwen & layer-matched & \textbf{11.72} & 13.5 & 0.989 & \textbf{0.011} & 0.067 (0.143) \\
Qwen & final-layer & 13.63 & 13.5 & 0.428 & 0.572 & 0.150 (0.143) \\
Llama & layer-matched & \textbf{13.67} & 15.5 & 0.979 & \textbf{0.021} & 0.073 (0.125) \\
Llama & final-layer & \textbf{13.09} & 15.5 & 0.998 & \textbf{0.002} & 0.099 (0.125) \\
\bottomrule
\end{tabular}
\caption{Layer distribution of bias selectivity. Mass-weighted centre of bias selectivity (\(\tilde S^{\text{unfair}} - \tilde S^{\text{fair}}\), positive part) against a permutation null over head positions, 20,000 draws (Monte-Carlo SE \(\leq 0.0035\)). \(p\)(late) tests concentration in late layers --- the direction the method reports; \(p\)(early) tests the opposite. Both references are shown because the original underspecifies \(\bar v_{\text{ref}}\) (Section~\ref{sec:contribution-score}); layer-matched is primary throughout this paper on reliability grounds (Table~\ref{tab:maps}). Values under the secondary labeller agree within Monte-Carlo error (Appendix~\ref{app:labeller-development-failure-audit}).}\label{tab:layers}
\end{table*}

\subsection{Score correlates within the terse arm}\label{app:terse-arm-decomposition}

Full decomposition behind
Section~\ref{sec:testing-the-length-hypothesis-an-inter}. Zero-order
correlations of the prescribed-set score within the terse arm: with the
fairness label \(+0.690\), with protected-group-token presence \(+0.671\), with
group-token count \(+0.485\), with response length \(-0.025\). Variance
explained: length alone 0.010, length plus group-token 0.534, length plus label
0.550, all three 0.558. Across all four arms, length and label together explain
0.519 and adding arm membership raises this to 0.783
(\(\Delta R^2 = +0.264\)). The increments between the label and group-token
models (\(+0.024\) and \(+0.008\)) and the corresponding partial correlations
are not interpretable at \(\phi = 0.951\), and we do not report them as
evidence.

\subsection{Unfairness by elicitation}\label{app:unfairness-by-elicitation}

\begin{table}[t]
\centering
\small
\begin{tabular}{@{}lll@{}}
\toprule
elicitation & Qwen & Llama \\
\midrule
Direct-Answer & 48/60 = 0.800 & 42/54 = 0.778 \\
Chain-of-Thought & 9/40 = 0.225 & 6/36 = 0.167 \\
gap & \textbf{0.575} & \textbf{0.611} \\
\bottomrule
\end{tabular}
\caption{Unfairness rate by elicitation, identification set.}\label{tab:elicit}
\end{table}

Directional error rates in Table~\ref{tab:labeller} are conditioned on the human label: false-unfair over the 25 human-fair responses, false-fair over the 91 human-unfair. Agreement by arm type: baseline 0.900, prescribed 1.000, complement 1.000, random-ablation 0.833 on \(n = 90\), with every interval overlapping the pooled interval.

\subsection{Control-map construction and reliability}\label{app:control-map-construction}

We construct two control maps from the same per-example scores: a \textbf{length
map} contrasting responses above and below median length while ignoring
fairness labels, and a \textbf{style map} contrasting DA and
CoT responses while ignoring labels. If the selection statistic
measures bias, the fairness map should be distinguishable from both.

\textbf{Score maps are reliably estimated.} Split-half reliability of the
fair/unfair gap map is 0.950 (Qwen) and 0.959 (Llama) under the layer-matched
reference, 0.907 and 0.913 under the final-layer one. Under label permutation
reliability collapses to approximately zero (0.001 DA, 0.009 CoT, -0.037 pooled
on Qwen), so low correlations between maps cannot be attributed to sampling
noise.

On Qwen under the study persona, nine of the ten highest-scoring heads in the
fairness map are among the ten most length-suppressed heads in the model.

\textbf{It is not the persona.} The persona alone shifts the outcome measure by
0.787 (Section~\ref{sec:the-outcome-measure-is-not-saturated}), so one might
attribute the correlation to it. Regenerating the identification set under
three persona strengths shows the association \emph{strengthens} as the persona
weakens. Under a neutral prompt, response length alone separates fair from
unfair responses at AUROC 0.996 --- on five unfair responses.

With the persona ladder above, four controls in total agree with the length reading. Every one is correlational, computed on data where the label and response length are close to inseparable, and Section~\ref{sec:testing-the-length-hypothesis-an-inter} rejects the hypothesis they support.

\subsection{Compute and software}\label{app:compute}

No model was trained or fine-tuned. All experiments are inference only: greedy
decoding at \texttt{max\_new\ =\ 128}, two instruction-tuned models of 7B and 8B
parameters, loaded in 4-bit NF4 \citep{dettmers2023qlora} through Hugging Face
Transformers \citep{wolf2020transformers} with the output projection
dequantised for scoring. The study comprises roughly 2{,}000 generations across
identification, the four-arm length intervention and all evaluation arms on
both models; every bootstrap, re-split and labeller re-derivation reported here
is arithmetic over stored score arrays and required no further model
evaluation.

All experiments ran on a single NVIDIA T4 (Google Colab free tier). GPU time was not logged; we estimate the total at under 40 GPU-hours, which includes two masking runs that were voided and regenerated (Appendix~\ref{app:reproducibility-notes}). Package versions: \texttt{transformers} 5.18.0, \texttt{bitsandbytes} 0.50.2, \texttt{torch} 2.11.0+cu130.

\end{document}